\documentclass[11pt]{article}

\usepackage[margin=1in]{geometry}
\usepackage[T1]{fontenc}
\usepackage[utf8]{inputenc}
\usepackage{lmodern}
\usepackage{microtype}
\usepackage{amsmath}
\usepackage{amssymb}
\usepackage{array}
\usepackage{booktabs}
\usepackage{graphicx}
\usepackage{subcaption}
\usepackage{float}
\usepackage{xcolor}
\usepackage{siunitx}
\usepackage{orcidlink}
\usepackage{hyperref}
\usepackage[nameinlink,noabbrev]{cleveref}

\graphicspath{{figures/}}

\hypersetup{
  colorlinks=true,
  linkcolor=blue,
  citecolor=blue,
  urlcolor=blue
}

\usepackage{mdframed}

\usepackage{pdfpages}

\title{When a better baseline makes delta learning worse: separating residual scale from learnability to guide target design}
\author{
  \orcidlink{0000-0002-1168-9830}\hspace{1mm}Kareem M.~Gameel\\
  Department of Physical and Environmental Sciences, University of Toronto Scarborough\\
  Alliance for AI-Accelerated Materials Discovery (A3MD)\\
  Toronto, Ontario, Canada
  \and
  \orcidlink{0000-0001-9906-8010}\hspace{1mm}Ihor Neporozhnii\\
  Department of Physical and Environmental Sciences, University of Toronto Scarborough\\
  Alliance for AI-Accelerated Materials Discovery (A3MD)\\
  Toronto, Ontario, Canada
  \and
  \orcidlink{0000-0002-3099-585X}\hspace{1mm}Sjoerd Hoogland\\
  Alliance for AI-Accelerated Materials Discovery (A3MD)\\
  Department of Electrical and Computer Engineering, University of Toronto\\
  Toronto, Ontario, Canada
  \and
  \orcidlink{0000-0002-8656-5074}\hspace{1mm}Oleksandr Voznyy\\
  Department of Chemistry, University of Toronto\\
  Department of Physical and Environmental Sciences, University of Toronto Scarborough\\
  Alliance for AI-Accelerated Materials Discovery (A3MD)\\
  Toronto, Ontario, Canada
}
\date{}

\begin{document}

\maketitle

\begin{abstract}
In scientific machine learning, $\Delta$-learning trains models on residual errors relative to approximate baselines, assuming that smaller residual scales inherently yield better downstream accuracy. Here, we demonstrate that residual scale alone is an insufficient heuristic for learnability. For molecular total energies, replacing a linear baseline with a neural network fitted to the same elemental counts reduced the residual scale, yet more than doubled GemNet’s prediction error. Across local structural descriptors, more flexible baselines produced smaller residuals that varied sharply between structurally similar molecules. By contrast, semi-empirical baselines left larger residuals, yet yielded lower test error and improved out-of-domain transferability. Across the targets studied, scale-normalized graph Dirichlet roughness tracked scale-normalized prediction error, suggesting its use as a pre-training diagnostic for target selection. These results motivate \emph{baseline complementarity} as a target-design principle in scientific machine learning: a useful baseline captures contributions the downstream architecture struggles to predict.
\end{abstract}

\section{Introduction}
\label{sec:introduction}

Supervised machine learning learns a mapping from an input representation to a target property. The generalizability of this mapping fundamentally depends on its smoothness—the degree to which continuous, localized shifts in feature space produce smooth variations in the target. Conventional model development approaches this problem primarily from the representation side, refining neural architectures and inductive biases to construct feature spaces where the target varies smoothly \cite{behler2007generalized,bartok2010gap,gasteiger2022gemnetoc,schutt2023schnetpack2,batzner2022nequip}. The prediction target itself, by contrast, is usually treated as fixed.

Yet heterogeneous targets, such as molecular total energies, aggregate compositional, local atomic, many-body, and long-range electronic interactions that span diverse physical scales. A given graph neural network (GNN) architecture may represent these distinct physical contributions with varying fidelity. When a component of the target aligns poorly with a model's intrinsic representation, it increases target roughness within that feature space, placing a fundamental bottleneck on downstream generalizability.

$\Delta$-learning is conventionally motivated by a simple heuristic: subtracting a more accurate physical baseline leaves a smaller residual, thereby easing the downstream prediction task \cite{ramakrishnan2015delta,chen2023solutiondelta,dral2020hierarchical,grumet2024dielectric,chang2025activation}. In practice, however, this assumption does not always hold \cite{atz2022deltaqml}. For example, in a study of isoelectronic diatomics, Krug \emph{et al.}~\cite{krug2025aha} demonstrated that while their proposed baseline reduced the training data required for chemical accuracy by up to an order of magnitude, an alternative baseline combining an alchemical harmonic approximation with a Morse potential yielded negligible gains—and occasionally worsened predictions compared to direct target learning. These contrasting outcomes expose a central, unanswered question in scientific AI: \emph{what does subtracting a baseline alter about the learning task beyond simply shrinking the target's scale?}

Subtracting a baseline simultaneously alters two distinct properties of the target: its \textbf{scale} and its \textbf{functional form}. Holding scale-normalized learnability constant, a smaller residual directly yields a lower absolute prediction error. However, altering the functional form can cause the residual to vary more sharply between inputs that the network represents similarly—effectively increasing prediction error relative to target scale. $\Delta$-learning must therefore be reframed as an active problem of \emph{target design}, where candidate baselines are evaluated by their dual ability to compress target scale while preserving or enhancing smoothness within the downstream architecture's feature space.

To decouple intrinsic learnability from target scale post-training, we define the scale-normalized mean absolute error ($n\text{-MAE}$) as the prediction MAE divided by the target interquartile range ($\text{IQR}$). An increase in $n\text{-MAE}$ reveals that a residual target has become intrinsically harder for the architecture to resolve relative to its spread. The ideal baseline transformation reduces overall target scale without incurring an $n\text{-MAE}$ penalty.

Because $n\text{-MAE}$ can only be measured after model training, rational target design requires a pre-training diagnostic of feature-space smoothness~\cite{aldeghi2022roughness,graff2023rogiXD}. Here, we introduce IQR-normalized graph Dirichlet roughness ($D_{\rm IQR}$), a metric that directly quantifies target variation across fixed, architecture-informed feature spaces prior to training \cite{belkin2006manifold}. A low $D_{\rm IQR}$ value indicates that molecules positioned close together in feature space exhibit consistent target values relative to scale, providing an \emph{a priori} screening tool for candidate baselines.

We systematically evaluate this target-design framework across molecular total energy predictions using GemNet-T and SchNet \cite{gasteiger2021gemnet,schutt2023schnetpack2}, comparing linear descriptors, non-linear machine-learned fits, and two semi-empirical quantum-mechanical baselines, GFN2-xTB \cite{bannwarth2019gfn2xtb} and density-functional tight binding (DFTB). In-domain accuracy is benchmarked on organometallic complexes from tmQM, and out-of-distribution transferability is tested on organic molecules from QM9 \cite{balcells2020tmqm,ramakrishnan2014quantum}. Across both architectures, replacing simple linear baselines with flexible non-linear fits yields smaller but rougher residuals and higher normalized error. The xTB baseline reduces target scale and normalized roughness and improves both in-domain accuracy and out-of-domain transferability.

Ultimately, these findings frame $\Delta$-learning as a two-dimensional optimization problem over \textbf{residual scale} and \textbf{representation-space smoothness}, suggesting that target formulation deserves consideration alongside model architecture in scientific machine learning.

\section{Results}
\label{sec:results}

\subsection{Baseline flexibility at fixed information}
\label{sec:reference-correction}

To test whether minimizing residual scale inherently improves downstream accuracy, we compare baselines that differ in flexibility but use the same elemental counts. Following standard practice \cite{tran2023oc22,schutt2023schnetpack2,pelaez2024torchmdnet2}, we subtract fixed element-wise atomic references from the total DFT energy to obtain the reference-corrected energy ($E_{\rm ref}$). We then fit a linear Bag-of-Elements (BoE) model to atomic counts and subtract its prediction ($E_{\rm BoE}$) to construct the residual target:

\begin{equation}
E_{\rm ref,BoE} = E_{\rm ref} - E_{\rm BoE}
\label{eq:boe-reference-correction}
\end{equation}

Because the subtracted baseline is restored at inference during stacked $\Delta$-learning, the mean absolute error (MAE) of the residual prediction is identically equal to that of the complete pipeline (SI Section 1).

\begin{figure}[H]
  \centering
  \includegraphics[width=\linewidth]{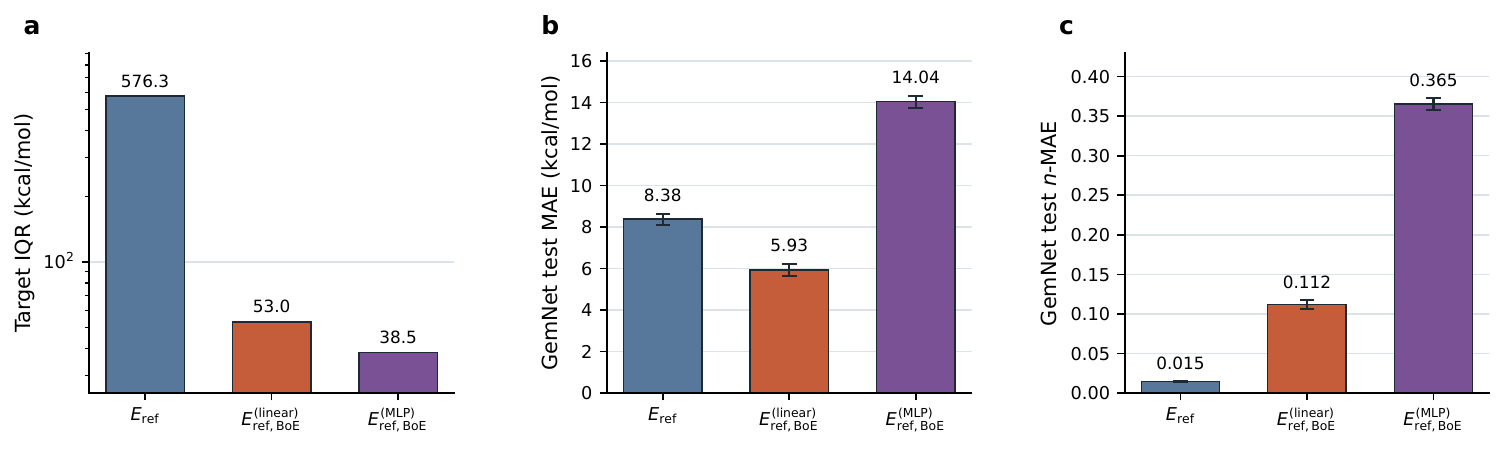}
  \caption{\textbf{Target scale and GemNet accuracy under linear and non-linear BoE corrections.} \textbf{a,} Test-set interquartile range (IQR) for the reference energy $E_{\rm ref}$ and its linear- and MLP-BoE residuals on tmQM. \textbf{b,} Mean GemNet test MAE across targets. \textbf{c,} Scale-normalized test error ($n\text{-MAE}_{\rm test}$). Error bars represent sample standard deviations across three independent training seeds.}
  \label{fig:reference-correction}
\end{figure}

We report the test mean absolute error ($\operatorname{MAE}_{\rm test}$) and its scale-normalized counterpart,

\begin{equation}
n\text{-MAE}_{\rm test} = \frac{\operatorname{MAE}_{\rm test}}{\operatorname{IQR}_{\rm test}(y)}
\end{equation}

The normalized error expresses the prediction error relative to the spread of the test target and is invariant to changes in energy units (SI Note 2).

Evaluating GemNet-T on the tmQM dataset reveals a trade-off between target scale and learnability (Fig.~\ref{fig:reference-correction}). Applying the linear BoE correction compresses the target IQR by a factor of 10.9, but increases GemNet’s normalized error by a factor of 7.7 (Fig.~\ref{fig:reference-correction}a,c). Although the residual becomes harder for GemNet to learn relative to its spread, the reduction in scale outweighs the increase in normalized error, yielding a 1.41-fold net reduction in absolute test MAE (Fig.~\ref{fig:reference-correction}b).

This gain in absolute accuracy motivates a question: \emph{does pushing residual scale even smaller further improve predictions?} To test this without injecting extra structural information, we replace the linear baseline with a more flexible multilayer perceptron (MLP) trained on the same elemental counts. The MLP baseline reduces the residual spread by an additional factor of 1.38 relative to the linear model. However, GemNet’s normalized error increases by a factor of 3.27 (Fig.~\ref{fig:reference-correction}a,c). The increase in normalized error outweighs the reduction in target scale, causing GemNet’s final test MAE to more than double (2.37$\times$) compared to the linear baseline (Fig.~\ref{fig:reference-correction}b). SchNet shows the same trend (SI Section 1). Across seeds, both architectures achieved lower mean final-epoch training MAE on the MLP-BoE residual than on the linear-BoE residual, yet their best validation and corresponding test MAEs were higher (Supplementary Table~S3).

These results demonstrate that residual scale alone does not predict downstream accuracy. Because normalized error can only be measured post-training, selecting effective baselines requires an \emph{a priori} diagnostic of target learnability. To evaluate candidate targets prior to model training, we next investigate how smoothly residual values vary across molecular structural representations.

\subsection{Pre-training roughness tracks target learnability \emph{a priori}}
\label{sec:pretraining-roughness}

To evaluate candidate residuals before downstream training, we measure how target values vary across fixed geometric feature spaces built to serve as \textbf{representation proxies} for the downstream architectures. The \textbf{Pair-Radial Expansion (PRE)} describes molecular geometry through pairwise distance distributions, acting as a proxy for SchNet’s pairwise representation. \textbf{PRE+A} extends PRE by incorporating central-element triplet angle distributions, capturing the directional and angular information accessible to GemNet-T. (Both proxy spaces are later reused as baseline models.)

Within each proxy space, we construct a $k$-nearest-neighbor graph connecting test-set molecules with similar structural representations and compute the scale-normalized graph Dirichlet roughness, $D_{\rm IQR}(y)$ (Methods, Section~\ref{sec:methods-roughness-diagnostics}, Equation~\ref{eq:dirichlet-roughness}). This metric measures target variation between neighboring molecules relative to the central target spread ($\text{IQR}$), where lower scores indicate that geometrically similar molecules possess more consistent target values. Applied to the composition-corrected targets, $D_{\rm IQR}$ resolves the learnability paradox from Figure~\ref{fig:reference-correction}: in the GemNet-T proxy space (PRE+A), normalized roughness increases from $0.548$ for $E_{\rm ref}$ to $1.127$ for the linear-BoE residual, surging to $1.658$ for the MLP-BoE residual (and $1.709$ vs $1.152$ in the SchNet proxy space, PRE; SI Note~6). Pre-training target roughness thus directly tracks GemNet-T’s post-training normalized error ($n\text{-MAE}_{\rm test}$), demonstrating that non-linear composition baselines shrink target scale only by injecting fine-grained roughness into the target landscape.

\subsection{Residual roughness guides target design beyond baseline accuracy}
\label{sec:residual-design}

For baselines fitted to elemental counts, normalized roughness tracked GemNet’s error relative to target scale. We next test whether this relationship extends to baselines that incorporate molecular structure. We fit linear and MLP baselines to three feature families of increasing structural detail (Table~\ref{tab:baseline-models}): the Bag-of-Bonds (BoB) descriptor (counting bonds by type), along with PRE and PRE+A—the geometric representation proxies introduced in §\ref{sec:pretraining-roughness}, which now serve directly as inputs to the baseline models.

\begin{table}[H]
  \centering
  \caption{\textbf{Baseline models used for target transformation.}}
  \label{tab:baseline-models}
  \small
  \setlength{\tabcolsep}{4pt}
  \renewcommand{\arraystretch}{1.2}
  \begin{tabular}{@{}>{\raggedright\arraybackslash}p{0.12\linewidth}>{\raggedright\arraybackslash}p{0.25\linewidth}>{\raggedright\arraybackslash}p{0.19\linewidth}>{\raggedright\arraybackslash}p{0.38\linewidth}@{}}
    \toprule
    Baseline model & Input features & Fitting approach & Physical/representation target \\
    \midrule
    BoB & RDKit bond-type counts rather than conventional BoB Coulomb-matrix entries & Linear / MLP & Compositional and functional-group variation \\
    PRE & Pairwise distances (\(r_{ij}\)) & Linear / MLP & Local two-body geometry (SchNet proxy) \\
    PRE+A & Distances and angles (\(r_{ij},\theta_{ijk}\)) & Linear / MLP & Local three-body geometry (GemNet proxy) \\
    xTB & Atomic identities and molecular coordinates & Semiempirical QM & Non-local electronic and higher-order interactions \\
    DFTB\(\dagger\) & Atomic identities and molecular coordinates & Semiempirical QM & Molecule-wide electronic structure; supporting comparison with xTB \\
    \bottomrule
  \end{tabular}
\end{table}

All three descriptor families use the same target: the atom- and composition-referenced DFT energy, $E_{\rm ref,BoE}^{\rm DFT}$. For each fitted baseline $b$, the downstream GNN learns the residual difference:

\begin{equation}
\Delta^b=E_{\rm ref,BoE}^{\rm DFT}-\widehat{E}_b,
\label{eq:descriptor-delta}
\end{equation}

This experimental setup separates two distinct effects. Comparing BoB, PRE, and PRE+A at a fixed MLP architecture tests what the baseline gains from additional structural information; comparing linear and MLP fits within the same descriptor family tests the effect of baseline flexibility while holding input information constant.

For the physical baseline comparison, we use the energy from a semi-empirical xTB calculation \cite{bannwarth2019gfn2xtb}, applying the same referencing protocol before taking the residual difference:

\begin{equation}
\Delta^{\rm xTB}
=E_{\rm ref,BoE}^{\rm DFT}-E_{\rm ref,BoE}^{\rm xTB}.
\label{eq:xtb-delta}
\end{equation}

\begin{figure}[t]
  \centering
  \includegraphics[width=\linewidth]{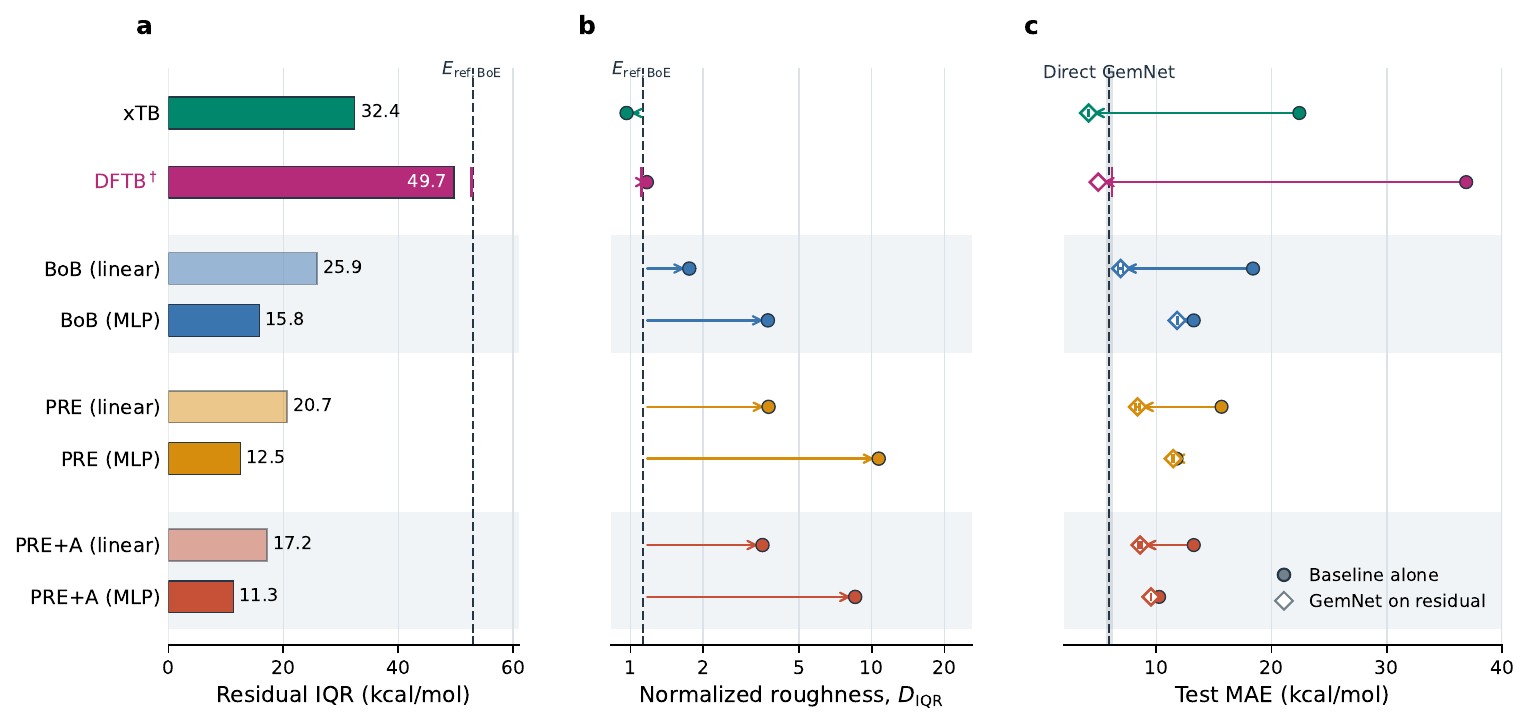}
  \caption{\textbf{Baseline accuracy, residual roughness, and GemNet test performance.} \textbf{a,} Test residual IQR. \textbf{b,} Normalized roughness on the test PRE+A graph (log scale). \textbf{c,} Baseline-only test MAE (filled circles) and GemNet residual test MAE (open diamonds). Dashed lines mark the DFT-BoE parent or direct-GemNet reference; arrows show the corresponding changes. Main GNN results are three-seed means; error bars and shading show sample SD. Magenta DFTB\(^\dagger\) is a single-seed, valid-subset control, excluded from pooled analyses. Short magenta ticks mark its matched-subset parent references (SI Note~8).}
  \label{fig:residual-design}
\end{figure}

Adding structural detail reveals the same flexibility penalty. As expected, incorporating structural information improves the MLPs' native predictions: baseline error falls from BoB to PRE to PRE+A, producing residuals narrower than the DFT parent by factors of 3.3, 4.2, and 4.7 in test IQR (Fig.~\ref{fig:residual-design}a,c). Yet all three residuals vary more sharply between molecules that are close in the GemNet proxy space (Fig.~\ref{fig:residual-design}b), and none improves GemNet’s test MAE over direct training on the DFT parent (Fig.~\ref{fig:residual-design}c). Thus, the descriptor baseline that best approximates DFT energies leaves the smallest target, but that target does not yield the most accurate final prediction.

The same pattern holds when descriptor complexity is held fixed. Within each family, replacing the linear fit with an MLP produces a smaller, rougher residual and increases GemNet’s test error. For example, comparing the two BoB baselines, the MLP reduces the residual IQR by a further factor of 1.6, yet GemNet’s test MAE rises by a factor of 1.7 (Fig.~\ref{fig:residual-design}). The trade-off observed with elemental counts in §\ref{sec:reference-correction} therefore persists across structured representations.

By contrast, the semi-empirical xTB baseline yields higher accuracy. Although xTB leaves a larger residual than any of the three descriptor-based MLPs, GemNet predicts it more accurately. Relative to the DFT parent, normalized roughness falls from 1.127 to 0.964, and mean GemNet test MAE drops from 5.93 to 4.14~kcal/mol (Fig.~\ref{fig:residual-design}b,c). This advantage persists after accounting for target scale: GemNet’s normalized test error is 7.2 times higher on the MLP-PRE residual than on the xTB residual (and 4.9 times higher for SchNet; SI Note~6). These results align with baseline complementarity—akin to observations by Tu and Rowley \cite{tu2026transferable}, where a DFTB3 baseline supplied long-range electrostatic contributions beyond a local network’s cutoff to improve interaction energies.

Finally, the DFTB control \cite{elstner1998sccdftb,hourahine2020dftbplus} illustrates why absolute accuracy and normalized roughness must still be distinguished. Because DFTB calculations failed for 362 neutral tmQM molecules, we evaluated one GNN seed per architecture on the successful subset (excluding this control from pooled analyses; SI Note~8). On that matched subset, GemNet test MAE fell from 6.15 to 4.98~kcal/mol even though normalized roughness in PRE+A space rose from 1.113 to 1.169. While lower roughness accompanied the accuracy gains for xTB, the DFTB comparison shows that absolute error can improve even when normalized roughness increases.

\subsection{Model-relative target roughness tracks scale-normalized learnability}
\label{sec:roughness-predicts}

\begin{figure}[H]
  \centering
  \includegraphics[width=0.90\linewidth]{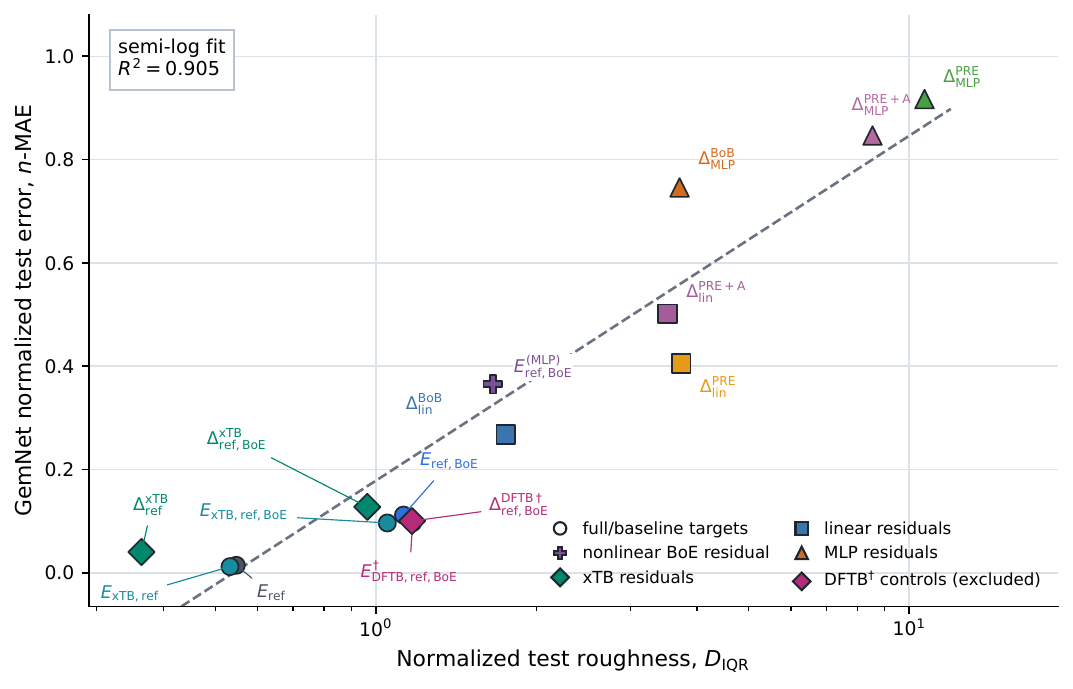}
  \caption{\textbf{Model-relative roughness and normalized GemNet test error.} Mean GemNet test \(n\)-MAE, defined as \(\operatorname{MAE}_{\rm test}/\operatorname{IQR}_{\rm test}(y)\), is plotted against test \(D_{\rm IQR}\) for 13 primary targets. Roughness is measured on the standardized test PRE+A graph, and the horizontal axis is logarithmic. Marker shapes identify the target classes shown in the legend. Points show means across three GNN seeds. The dashed line is the semi-log regression over the 13 primary target means (\(R^2=0.905\)). Magenta DFTB\(^\dagger\) points are single-seed, valid-subset controls, excluded from the regression and correlations.}
  \label{fig:roughness-nmae}
\end{figure}

Because $D_{\rm IQR}$ and $n\text{-MAE}$ share target IQR as a scaling factor, part of their strong correlation reflects shared-denominator normalization~\cite{pearson1897spurious}; nevertheless, joint-label permutation tests confirm that architecture-aligned target roughness tracks relative learnability beyond this scaling effect. Across all 13 primary targets, GemNet’s scale-normalized test error ($n\text{-MAE}_{\rm test}$) increases approximately linearly with log PRE+A Dirichlet roughness, yielding $R^2 = 0.905$ (Fig.~\ref{fig:roughness-nmae}; SchNet exhibits $R^2 = 0.946$ in PRE space; SI Note~6). While unnormalized metrics show no such trend (SI Note~7), shuffling target values among molecules preserves target IQRs while destroying structural alignment; across 5,000 joint-label permutations, none produced a positive correlation as high as the original molecular assignment (median shuffled $R^2 = 0.862$; one-sided $p \approx 0.0002$~\cite{phipson2010permutation}; SI Note~7). These retrospective results support $D_{\rm IQR}$ as a candidate scale-invariant pre-training diagnostic of relative target learnability. Its ability to guide baseline selection for new target families remains to be tested.

\subsection{Target design governs out-of-domain transferability}
\label{sec:ood-transfer}

To test whether baseline choice affects transfer across chemical domains, we evaluated models trained on tmQM transition-metal complexes using small organic molecules from QM9 \cite{ramakrishnan2014quantum}. We recomputed QM9 reference energies at the identical level of DFT theory and kept all baseline parameters fitted on tmQM fixed. Because out-of-domain (OOD) data reside in a shifted energy manifold, models inevitably suffer from global miscalibration, rendering absolute error metrics like raw MAE sensitive to dataset-wide shifts rather than local predictive fidelity. Because relative energy ordering and potential energy surface topography govern chemical phenomena such as structural stability and reaction energetics, trend preservation is essential for assessing whether a model has learned transferable chemical physics rather than domain-specific offsets. To rigorously assess this trend preservation, we evaluate model predictions using shift-independent metrics: Kendall rank correlation ($\tau$) to measure energetic ordering regardless of global shifts, and mean-shifted MAE to quantify residual scatter after removing a single dataset-wide offset estimated from QM9 reference labels (SI Note~5).

\begin{figure}[H]
  \centering
  \includegraphics[width=\linewidth]{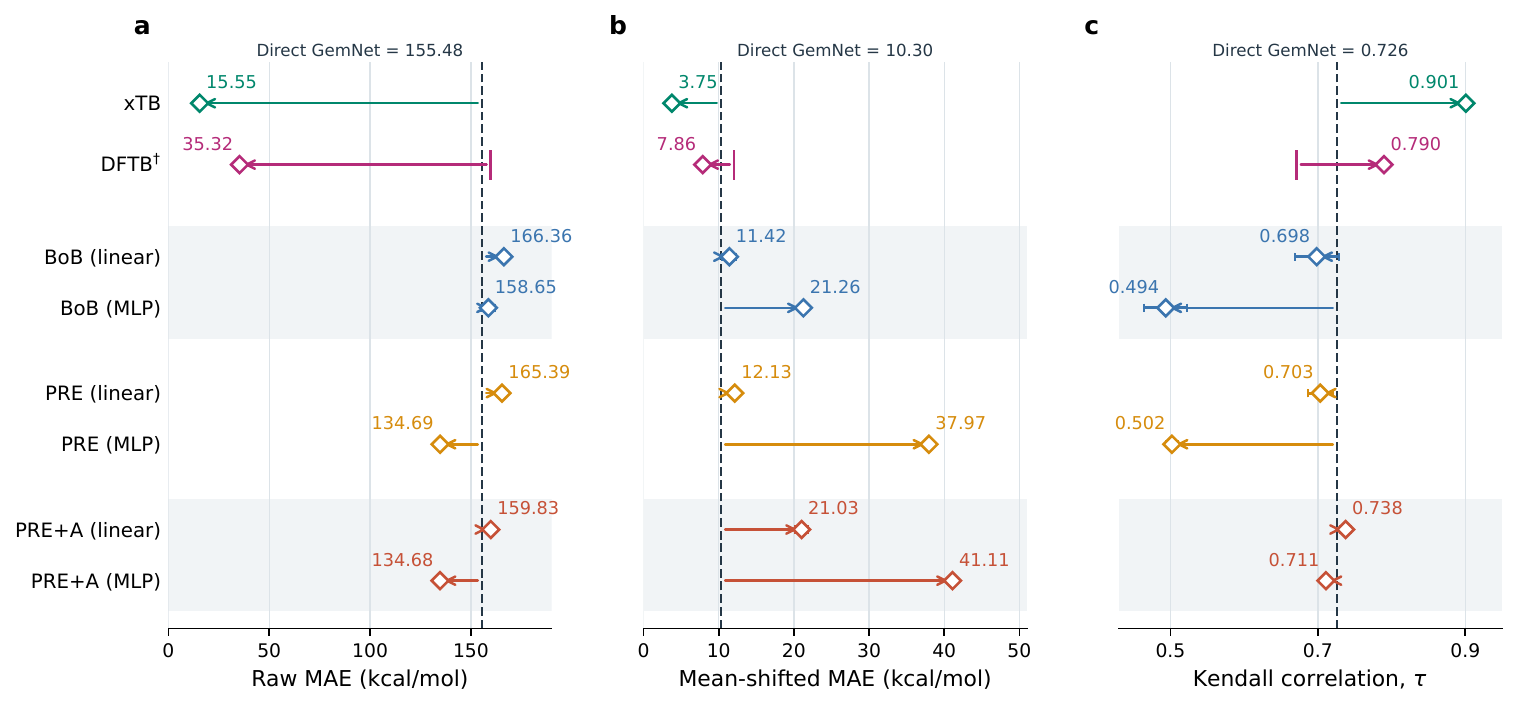}
  \caption{\textbf{Out-of-domain GemNet transfer from tmQM to QM9.} \textbf{a,} Raw MAE. \textbf{b,} Mean-shifted MAE using one offset estimated from QM9 labels. \textbf{c,} Kendall correlation. Diamonds show baseline-plus-GemNet predictions against \(E_{\rm ref,BoE}\); dashed lines mark direct GemNet. Primary results are three-seed means with sample-SD bars on 128,988 common molecules. Magenta DFTB\(^\dagger\) is a seed-101 control on 128,978 valid molecules; its short reference ticks and arrows use direct seed 101 on the same subset (SI Notes~5 and 8).}
  \label{fig:qm9-transfer}
\end{figure}

On the common QM9 cohort, standalone xTB achieved a raw MAE of $16.73$~kcal/mol, a mean-shifted MAE of $11.35$~kcal/mol, and Kendall $\tau = 0.677$. Adding the GemNet correction reduced raw MAE modestly to $15.55$~kcal/mol, but lowered mean-shifted MAE to $3.75$~kcal/mol and raised $\tau$ to $0.901$ (Fig.~\ref{fig:qm9-transfer}). Direct GemNet further illustrates why these metrics are necessary: its raw MAE was $155.48$~kcal/mol—more than nine times that of xTB—yet its higher Kendall score ($\tau = 0.726$) shows that GemNet captured underlying chemical trends better but suffered from a severe global offset. Correcting for this offset reveals a mean-shifted MAE of $10.30$~kcal/mol, outperforming standalone xTB. Thus, xTB provided baseline calibration, while the learned correction substantially refined molecular energy ordering and reduced residual scatter. SchNet exhibited parallel improvements, raising Kendall $\tau$ from $0.746$ (direct learning) to $0.895$ (xTB residual; SI Note~5).

By contrast, descriptor-based MLPs failed to improve transferability. No local descriptor baseline surpassed direct GemNet across both trend metrics; for example, linear PRE+A increased mean-shifted MAE to $21.03$~kcal/mol despite a minor gain in Kendall $\tau$ ($0.738$). DFTB provided a second test of electronic-structure baseline complementarity: on its valid subset, DFTB + GemNet reduced raw MAE from $159.84$ to $35.32$~kcal/mol, lowered mean-shifted MAE from $12.03$ to $7.86$~kcal/mol, and increased Kendall $\tau$ from $0.671$ to $0.790$ relative to direct GemNet (Fig.~\ref{fig:qm9-transfer}; SI Notes~5 and 8).

These OOD results reinforce the principle of baseline complementarity. Local geometric descriptors mirror message-passing GNN inductive biases, stripping away smooth local features and yielding rough, difficult targets. Conversely, semi-empirical quantum methods capture long-range electrostatics and charge redistribution that local networks represent poorly. By preserving or enhancing scale-normalized representation smoothness ($D_{\rm IQR}$) while shrinking target scale, complementary baselines systematically support both in-domain generalization and out-of-domain transferability.

\section{Discussion}
\label{sec:discussion}

In scientific machine learning, $\Delta$-learning rests on a compelling intuition: rather than predicting a high-fidelity quantum-mechanical energy from scratch, a GNN should learn the correction to a cheap approximate baseline~\cite{zheng2021aiqm1,zeng2023qdpi}. Because the baseline captures much of the variation in the target, the resulting residual is smaller and presumed easier to learn. Yet, as our findings demonstrate, this intuition does not always hold. Aggressively shrinking total target scale does not guarantee downstream learnability, explaining why $\Delta$-learning has often relied on trial-and-error heuristics rather than design principles.

Our results resolve this inconsistency by reframing $\Delta$-learning as a functional target transformation governed by two distinct properties: \textbf{target scale} and \textbf{scale-normalized representation smoothness}, which tracks downstream learnability as assessed by $n\text{-MAE}$. Because absolute test error is the product of target scale and normalized error, aggressive scale reduction can still yield a net accuracy gain if the shrinkage outweighs the accompanying loss in learnability—as seen with the linear Bag-of-Elements (BoE) baseline. However, pushing shrinkage even further with a nonlinear MLP-BoE baseline raises normalized error enough to outweigh the further reduction in target scale and degrades performance.

Optimal $\Delta$-learning therefore requires \textbf{baseline complementarity}: we propose prioritizing baselines that capture the specific components responsible for the model's error, rather than merely absorbing the largest share of total target spread. Local descriptor baselines (Bag-of-Bonds, PRE, and PRE+A) share geometric information with local message-passing GNNs; they strip away smooth local features the network already represents efficiently, leaving residuals that are narrower in scale yet disproportionately rough and harder to learn, resulting in poorer test performance than direct learning. Conversely, semi-empirical quantum methods (xTB and DFTB) capture electronic effects—such as charge redistribution and long-range electrostatics~\cite{wengert2022cocrystals,tu2026transferable}—that may contribute to the errors of local architectures. By absorbing the specific energy contributions that drive GNN error, complementary baselines shrink target scale while preserving or reducing scale-normalized roughness, consistently improving in-domain test accuracy (tmQM) and enhancing transferability to out of distribution chemical domains (QM9).

Because post-training evaluation requires expensive model optimization, practical target engineering requires pre-training diagnostics. To address this, we introduced the scale-normalized graph Dirichlet roughness ($D_{\rm IQR}$), which tracks target variation over architecture-aligned representation proxies without training a downstream GNN. Its strong semi-log association with normalized test error ($n\text{-MAE}$) supports $D_{\rm IQR}$ as a candidate pre-training diagnostic for screening baselines. While shared normalization contributes substantially to this association, joint-label permutation controls support an additional contribution from molecular-neighbourhood alignment. Finally, our evaluation focuses on equilibrium molecular energies without force supervision; extending this target-design framework to joint energy-and-force training and non-equilibrium molecular dynamics simulations remains an important direction for future work.

\section{Conclusion}
\label{sec:conclusion}

This study advances $\Delta$-learning from heuristic baseline selection toward principled target design. Our results suggest that baseline complementarity and the balance between target shrinkage and residual smoothness shape downstream predictive accuracy. This positions target formulation as a foundational design choice alongside network architecture, with potential implications for residual learning, multi-fidelity modeling, and scientific machine learning surrogates.

\section{Methods}
\label{sec:methods}

\subsection{Datasets and targets}
\label{sec:methods-datasets-targets}

The in-domain data are the aligned neutral tmQM transition-metal complexes with \(\omega\)B97M-V/def2-SVPD energies and fixed train, validation, and test molecule identifiers \cite{balcells2020tmqm,mardirossian2016wb97mv,rappoport2010def2svpd}. The full panel contains 56,830 training, 7,104 validation, and 7,104 test molecules. The six descriptor residuals use 56,821 training and 7,103 validation molecules, while all 13 primary targets share the same 7,104 test molecules. Both architectures use identical target values and split memberships within each target. DFTB success filtering leaves 56,540 training, 7,068 validation, and 7,068 test molecules; failed calculations were not imputed. QM9 is used only for OOD evaluation; its energies were recomputed at the same level with Psi4 to avoid conflating chemical-domain shift with electronic-structure mismatch \cite{ramakrishnan2014quantum,parrish2017psi4}.

For method \(m\), molecular energies were first corrected using fixed element-wise atomic references,
\begin{equation}
  E_{\rm ref}^{m}(i)=E_{\rm raw}^{m}(i)-\sum_Z n_Z(i)\epsilon_Z^{m},
  \label{eq:atomic-reference}
\end{equation}
then corrected by a linear model of elemental counts fitted only on the tmQM training split to obtain \(E_{\rm ref,BoE}^{m}\). Atomic references are fixed external inputs, not parameters fitted to held-out molecular energies. Separate composition models are fitted for DFT, xTB, and DFTB and frozen for validation, test, and QM9 evaluation. The DFT quantity is the parent target. Delta targets subtract either a fitted descriptor baseline or the consistently referenced xTB/DFTB energy. We retained the split identifiers during target generation and aligned target tables, LMDB records, and predictions by molecule ID or explicitly mapped sample ID.

\subsection{Baseline models}
\label{sec:methods-baselines}

The linear BoE reference uses 44 elemental counts and an intercept. A composition-only control fits a nonlinear MLP to the same counts and atom-referenced DFT energy, thereby changing baseline flexibility without adding structural information. BoB denotes the study-specific RDKit-derived typed-bond-count vector, not the conventional Coulomb-matrix Bag-of-Bonds descriptor. PRE expands local element-resolved pair distances in Gaussian radial basis functions up to a spatial cutoff, and PRE+A augments PRE with central-element-resolved triplet angle distributions. BoB has 4,950 channels. PRE has 34,650 pair-radial features, 35 Gaussian channels, and a \(6.0~\text{\AA}\) cutoff. Centres span 0.8--\(3.0~\text{\AA}\) at \(0.1~\text{\AA}\) intervals (width \(0.12~\text{\AA}\)), then 3.25--\(6.0~\text{\AA}\) at \(0.25~\text{\AA}\) intervals (width \(0.30~\text{\AA}\)). PRE+A adds 11,088 angular features, giving 45,738 total: 44 central elements, 21 unordered pairs of six radial shells, and 12 \(\cos\theta\) bins.

BoB, PRE, and PRE+A each have linear ridge and MLP fits to the DFT-BoE parent target; all four MLP baselines share hidden widths of 512 and 128 and the same training protocol. For BoB, PRE, and PRE+A, the MLP takes a single descriptor vector for each molecule and directly predicts its DFT-BoE energy. Ridge regularization was selected by five-fold training-only cross-validation, with shuffled folds (seed 42), mean held-out MAE as the selection criterion, and preprocessing refitted within each fold. Linear descriptor fits remove training-zero-variance channels and divide by the training population SD without centring. The descriptor ridge grid contains 25 logarithmically spaced values from \(10^{-6}\) to \(10^6\), with an interior optimum required. Final baseline parameters and preprocessing were fitted using only the corresponding tmQM training split. The validation split was used to select each MLP checkpoint, while test labels were used only for final evaluation. Baseline fitting seed 101 defines the canonical residual targets; fitting-seed sensitivity is reported separately in SI Note~9. The baseline MLPs use SiLU activations, dropout 0.10, and a scalar linear output. AdamW uses a learning rate of \(10^{-3}\), weight decay \(10^{-4}\), SmoothL1 loss with \(\beta=0.10\), batch size 1024, and gradient clipping at 1. Training runs for up to 800 epochs with early-stopping patience 80 and a minimum improvement of \(6.2751\times10^{-3}\)~kcal/mol.

GFN2-xTB and DFTB are fixed semiempirical baselines \cite{bannwarth2019gfn2xtb,elstner1998sccdftb,hourahine2020dftbplus}. DFTB uses SCC-DFTB/DFTB2 with the PTBP Slater--Koster set, without third-order or dispersion corrections \cite{dftborg2025ptbp}, and only paired successful calculations; failures were not imputed. Alongside the composition-referenced xTB residual (Equation~\ref{eq:xtb-delta}), we evaluated the atom-referenced residual \(\Delta_{\rm ref}^{\rm xTB}=E_{\rm ref}^{\rm DFT}-E_{\rm ref}^{\rm xTB}\). SI Note~8 reports DFTB coverage.

\subsection{GNN training}
\label{sec:methods-gnn-training}

GemNet-T is the primary downstream model and SchNet is the architecture control \cite{gasteiger2021gemnet,schutt2023schnetpack2}. Models were trained on energies only, without forces or periodic boundary conditions, using fixed splits, a \(6.0~\text{\AA}\) cutoff, and a common FairChem/OCP-style protocol. Molecular graphs were constructed on the fly. SchNet uses 128 hidden channels, 128 filters, six interactions, 50 Gaussian functions, and sum readout \cite{schutt2023schnetpack2}. GemNet-T uses three blocks, atom and edge widths 512, triplet and bilinear-triplet widths 64, radial and angular embedding widths 16, 128 radial functions, seven spherical functions, at most 50 neighbours, SiLU, and extensive summation \cite{gasteiger2021gemnet}. It uses Gaussian radial functions, spherical harmonics, and a polynomial envelope with exponent five.

Every target was standardized by its own training-set mean and population standard deviation. Within each architecture, architecture settings, loss, AdamW configuration, initial learning rate, scheduler rule, batch size, and checkpoint criterion were held fixed across targets. AdamW with AMSGrad used an initial learning rate of \(5\times10^{-4}\), MSE energy loss, and an 80-epoch budget; batch sizes were 16 for SchNet and four for GemNet. Evaluation used the same batch sizes as training. We used exponential moving average decay 0.999, gradient clipping at 10, and validation every 5,000 steps. ReduceLROnPlateau responded independently to each validation trajectory, with factor 0.8 and patience three, producing target-specific learning-rate schedules. Each run's best-validation checkpoint was evaluated on both the tmQM test set and QM9. All 13 primary targets use independent GNN seeds 101, 102, and 103 for both architectures; the two DFTB targets use seed 101 only for both architectures, giving 82 runs in total. Reported primary results are means of seed-level metrics, not ensemble predictions, with sample standard deviations shown where applicable. Single-seed DFTB results have no across-seed uncertainty estimate. Exact seed-level results are reported in SI Note~9. For Supplementary Table~S3, we averaged logged training-step MAEs during epoch 80. Validation and test MAEs refer to the validation-selected checkpoint. Supplementary Table~S19 gives the target normalization values shared across seeds and architectures.

\subsection{Roughness and transfer diagnostics}
\label{sec:methods-roughness-diagnostics}

GemNet roughness uses standardized PRE+A features and SchNet roughness uses standardized PRE features. A cosine \(k=10\) nearest-neighbour graph was built on the 7,104-molecule tmQM test cohort. Zero-variance feature channels were removed, and retained channels were divided by the graph cohort's population standard deviation without centring. This unsupervised graph preprocessing is distinct from train-only baseline preprocessing. Unique undirected union edges were assigned similarity-based weights using the locally scaled Gaussian kernel in Equation~\ref{eq:dirichlet-edge-weight}.
\begin{equation}
  w_{ij}=\exp\!\left(-\frac{d_{ij}^{2}}{\sigma_i\sigma_j}\right),
  \label{eq:dirichlet-edge-weight}
\end{equation}
Here, \(d_{ij}\) is the cosine distance between standardized descriptor vectors for molecules \(i\) and \(j\), and \(\sigma_i\) is a local distance scale, defined as the distance from molecule \(i\) to its \(k\)-th nearest neighbour.

Using these edge weights, we quantify target variation through the IQR-normalized graph Dirichlet roughness, \(D_{\rm IQR}\) \cite{belkin2006manifold}:
\begin{equation}
  D_{\rm IQR}(y)=\frac{1}{\sum_{\{i,j\}\in\mathcal E}w_{ij}}
  \sum_{\{i,j\}\in\mathcal E}
  w_{ij}\frac{(y_i-y_j)^2}{\operatorname{IQR}_{\rm test}(y)^2}.
  \label{eq:dirichlet-roughness}
\end{equation}
Here, \(\mathcal E\) is the set of unique undirected graph edges. For targets with nonzero IQR, this formulation normalizes local squared target differences by the squared central target spread, making \(D_{\rm IQR}\) invariant to nonzero uniform target rescaling.

Both \(D_{\rm IQR}\) and \(n\text{-MAE}_{\rm test}\) use the corresponding test-target IQR; graphs and weights remain fixed across primary targets within each architecture. DFTB controls use separately constructed valid-subset test graphs and are excluded from primary regressions and pooled statistics.

Quartiles use linear interpolation, and variance normalization uses population variance. The 13 target means describe related transformations of the same molecules. Semi-log regressions use unweighted ordinary least squares across the 13 primary target means, with an intercept and \(\ln D_{\rm IQR}\) as predictor. Raw-error, raw-roughness, variance-normalization, partial-correlation, and leave-one-target-out sensitivity analyses are reported in SI Note~7. The same note reports 5,000 joint test-label permutations, which preserve each target's distribution and IQR while breaking molecular-neighbourhood alignment, and comparisons of target pairs whose IQRs differ by at most 10\%. Each permutation (random seed 20260914) used the same molecule reassignment across all targets and both architectures, keeping graphs and model errors fixed. We compared the observed signed Pearson correlation with the shuffled correlations in a one-sided test, adding one to both the exceedance count and the number of permutations. For the partial-correlation analysis, we regressed \(\ln D_{\rm IQR}\) and \(\ln n\text{-MAE}\) separately on \(\ln \mathrm{IQR}_{\rm test}\) with an intercept and correlated the residuals. Leave-one-target-out fits are influence checks, not cross-validated prediction scores. These test-set analyses are retrospective diagnostics performed after model selection; test labels were not used to fit baselines, schedule training, or select checkpoints.

\subsection{OOD transfer evaluation}
\label{sec:methods-transfer}

tmQM-trained models were evaluated without retraining on QM9 geometries, using frozen tmQM training-only baseline parameters. Residual predictions were combined with their corresponding baselines and mapped to the common DFT-BoE parent before full-target correlation and ranking metrics were calculated. Primary parent-energy comparisons use 128,988 common QM9 molecules and three GNN seeds. The DFTB residual is shown on its 128,978-molecule intersection with this cohort, alongside a matched-subset direct seed-101 reference; this matches evaluation molecules, not training cohorts. Mean-shifted MAE removes one scalar mean-error offset estimated separately for each predictor from the evaluated QM9 reference labels and is used only as a post hoc oracle diagnostic, not an independently calibrated prediction score. Figure~\ref{fig:qm9-transfer} reports GemNet raw and mean-shifted MAEs in kcal/mol, without IQR normalization, and Kendall correlations; exact coverage, offset magnitudes, and additional metrics are reported in SI Note~5. For the atom-referenced DFT and nonlinear BoE routes, we subtracted the frozen linear DFT-BoE term after reconstruction to recover the common parent. We audited prediction arrays by checking molecule identifiers, complete target coverage, split disjointness, split-file checksums, and checkpoint identity across tmQM and QM9, and recomputed all 164 native-target MAEs.

\section*{Data and code availability}
The code, processed target tables, split identifiers, evaluation summaries, figure source data, manuscript sources, and reproducibility checks for this work are available in release \texttt{v0.4.2} of the project repository \cite{gameel2026correctedRelease}. The release includes the Psi4 total electronic energies of 130,807 QM9 molecules, computed at \(\omega\)B97M-V/def2-SVPD and provided in kcal/mol and in the original Hartree units alongside the QM9 molecule identifiers. This dataset is also available on Zenodo~\cite{gameelQM9Psi4Dataset}. The original tmQM and QM9 datasets are available from their respective providers \cite{balcells2020tmqm,ramakrishnan2014quantum}. Raw structures, LMDB databases, model checkpoints, and full prediction arrays are not redistributed in this release; the included manifests document their provenance.

\bibliographystyle{unsrt}
\bibliography{bib/references}

\begin{thebibliography}{10}

\bibitem{behler2007generalized}
J{\"o}rg Behler and Michele Parrinello.
\newblock Generalized neural-network representation of high-dimensional
  potential-energy surfaces.
\newblock {\em Physical Review Letters}, 98(14):146401, 2007.

\bibitem{bartok2010gap}
Albert~P. Bart{\'o}k, Mike~C. Payne, Risi Kondor, and G{\'a}bor Cs{\'a}nyi.
\newblock Gaussian approximation potentials: The accuracy of quantum mechanics,
  without the electrons.
\newblock {\em Physical Review Letters}, 104(13):136403, 2010.

\bibitem{gasteiger2022gemnetoc}
Johannes Gasteiger, Muhammed Shuaibi, Anuroop Sriram, Stephan G{\"u}nnemann,
  Zachary Ulissi, C.~Lawrence Zitnick, and Abhishek Das.
\newblock {GemNet-OC}: Developing graph neural networks for large and diverse
  molecular simulation datasets.
\newblock {\em Transactions on Machine Learning Research}, 2022.

\bibitem{schutt2023schnetpack2}
Kristof~T. Sch{\"u}tt, Stefaan S.~P. Hessmann, Niklas W.~A. Gebauer, Jonas
  Lederer, and Michael Gastegger.
\newblock {SchNetPack} 2.0: A neural network toolbox for atomistic machine
  learning.
\newblock {\em Journal of Chemical Physics}, 158(14):144801, 2023.

\bibitem{batzner2022nequip}
Simon Batzner, Albert Musaelian, Lixin Sun, Mario Geiger, Jonathan~P. Mailoa,
  Mordechai Kornbluth, Nicola Molinari, Tess~E. Smidt, and Boris Kozinsky.
\newblock {E(3)}-equivariant graph neural networks for data-efficient and
  accurate interatomic potentials.
\newblock {\em Nature Communications}, 13:2453, 2022.

\bibitem{ramakrishnan2015delta}
Raghunathan Ramakrishnan, Pavlo~O. Dral, Matthias Rupp, and O.~Anatole von
  Lilienfeld.
\newblock Big data meets quantum chemistry approximations: The $\delta$-machine
  learning approach.
\newblock {\em Journal of Chemical Theory and Computation}, 11(5):2087--2096,
  2015.

\bibitem{chen2023solutiondelta}
Xu~Chen, Pinyuan Li, Eugen Hruska, and Fang Liu.
\newblock $\delta$-machine learning for quantum chemistry prediction of
  solution-phase molecular properties at the ground and excited states.
\newblock {\em Physical Chemistry Chemical Physics}, 25(19):13417--13428, 2023.

\bibitem{dral2020hierarchical}
Pavlo~O. Dral, Alec Owens, Alexey Dral, and G{\'a}bor Cs{\'a}nyi.
\newblock Hierarchical machine learning of potential energy surfaces.
\newblock {\em Journal of Chemical Physics}, 152(20):204110, 2020.

\bibitem{grumet2024dielectric}
Manuel Grumet, Clara von Scarpatetti, Tom{\'a}{\v{s}} Bu{\v{c}}ko, and David~A.
  Egger.
\newblock Delta machine learning for predicting dielectric properties and raman
  spectra.
\newblock {\em The Journal of Physical Chemistry C}, 128(15):6464--6470, 2024.

\bibitem{chang2025activation}
Han-Chung Chang, Ming-Hsuan Tsai, and Yi-Pei Li.
\newblock Enhancing activation energy predictions under data constraints using
  graph neural networks.
\newblock {\em Journal of Chemical Information and Modeling}, 65(3):1367--1377,
  2025.

\bibitem{atz2022deltaqml}
Kenneth Atz, Clemens Isert, Markus N.~A. B{\"o}cker, Jos{\'e} Jim{\'e}nez-Luna,
  and Gisbert Schneider.
\newblock $\delta$-quantum machine-learning for medicinal chemistry.
\newblock {\em Physical Chemistry Chemical Physics}, 24:10775--10783, 2022.

\bibitem{krug2025aha}
Simon~L{\'e}on Krug, Danish Khan, and O.~Anatole von Lilienfeld.
\newblock Alchemical harmonic approximation based potential for iso-electronic
  diatomics: Foundational baseline for $\delta$-machine learning.
\newblock {\em The Journal of Chemical Physics}, 162(4):044101, 2025.

\bibitem{aldeghi2022roughness}
Matteo Aldeghi, David~E. Graff, Nathan Frey, Joseph~A. Morrone, Edward~O.
  Pyzer-Knapp, Kirk~E. Jordan, and Connor~W. Coley.
\newblock Roughness of molecular property landscapes and its impact on
  modellability.
\newblock {\em Journal of Chemical Information and Modeling},
  62(19):4660--4671, September 2022.

\bibitem{graff2023rogiXD}
David~E. Graff, Edward~O. Pyzer-Knapp, Kirk~E. Jordan, Eugene~I. Shakhnovich,
  and Connor~W. Coley.
\newblock Evaluating the roughness of structure--property relationships using
  pretrained molecular representations.
\newblock {\em Digital Discovery}, 2(5):1452--1460, 2023.

\bibitem{belkin2006manifold}
Mikhail Belkin, Partha Niyogi, and Vikas Sindhwani.
\newblock Manifold regularization: A geometric framework for learning from
  labeled and unlabeled examples.
\newblock {\em Journal of Machine Learning Research}, 7:2399--2434, 2006.

\bibitem{gasteiger2021gemnet}
Johannes Gasteiger, Florian Becker, and Stephan G{\"u}nnemann.
\newblock {GemNet}: Universal directional graph neural networks for molecules.
\newblock In {\em Advances in Neural Information Processing Systems},
  volume~34, pages 6790--6802, 2021.

\bibitem{bannwarth2019gfn2xtb}
Christoph Bannwarth, Sebastian Ehlert, and Stefan Grimme.
\newblock {GFN2-xTB}: An accurate and broadly parametrized self-consistent
  tight-binding quantum chemical method with multipole electrostatics and
  density-dependent dispersion contributions.
\newblock {\em Journal of Chemical Theory and Computation}, 15(3):1652--1671,
  2019.

\bibitem{balcells2020tmqm}
David Balcells and Bastian~Bjerkem Skjelstad.
\newblock {tmQM} dataset: Quantum geometries and properties of 86k transition
  metal complexes.
\newblock {\em Journal of Chemical Information and Modeling},
  60(12):6135--6146, 2020.

\bibitem{ramakrishnan2014quantum}
Raghunathan Ramakrishnan, Pavlo~O. Dral, Matthias Rupp, and O.~Anatole von
  Lilienfeld.
\newblock Quantum chemistry structures and properties of 134 kilo molecules.
\newblock {\em Scientific Data}, 1:140022, 2014.

\bibitem{tran2023oc22}
Richard Tran, Janice Lan, Muhammed Shuaibi, Brandon~M. Wood, Siddharth Goyal,
  Abhishek Das, Javier Heras-Domingo, Adeesh Kolluru, Ammar Rizvi, Nima Shoghi,
  Anuroop Sriram, F{\'e}lix Therrien, Jehad Abed, Oleksandr Voznyy, Edward~H.
  Sargent, Zachary~W. Ulissi, and C.~Lawrence Zitnick.
\newblock The open catalyst 2022 ({OC22}) dataset and challenges for oxide
  electrocatalysts.
\newblock {\em ACS Catalysis}, 13(5):3066--3084, 2023.

\bibitem{pelaez2024torchmdnet2}
Raul~P. Pelaez, Guillem Simeon, Raimondas Galvelis, Antonio Mirarchi, Peter
  Eastman, Stefan Doerr, Philipp Th{\"o}lke, Thomas~E. Markland, and Gianni
  De~Fabritiis.
\newblock {TorchMD-Net} 2.0: Fast neural network potentials for molecular
  simulations.
\newblock {\em Journal of Chemical Theory and Computation}, 20(10):4076--4087,
  2024.

\bibitem{tu2026transferable}
Nguyen Thien~Phuc Tu and Christopher~N. Rowley.
\newblock {$\Delta$}-learning for transferable machine learning interatomic
  potentials.
\newblock {\em The Journal of Chemical Physics}, 165:054114, 2026.

\bibitem{elstner1998sccdftb}
Marcus Elstner, Dirk Porezag, Gerd Jungnickel, Jens Elsner, Michael Haugk,
  Thomas Frauenheim, Sandor Suhai, and Gotthard Seifert.
\newblock Self-consistent-charge density-functional tight-binding method for
  simulations of complex materials properties.
\newblock {\em Physical Review B}, 58(11):7260--7268, 1998.

\bibitem{hourahine2020dftbplus}
Ben Hourahine, Balint Aradi, Volker Blum, Francesca Bonafe, Alexander Buccheri,
  Carlos Camacho, Cesar Cevallos, Mael~Y. Deshaye, Traian Dumitric{\u{a}},
  Alejandro Dominguez, Sebastian Ehlert, Marcus Elstner, Toma~S. van~der Heide,
  Jan Hermann, Stephan Irle, Julian~J. Kranz, Carsten K{\"o}hler, Tim
  Kowalczyk, Tom{\'a}{\v{s}} Kuba{\v{r}}, I.-S. Lee, Vitali Lutsker,
  Reinhard~J. Maurer, Seung~Kyu Min, Iain Mitchell, Christian Negre, Thomas~A.
  Niehaus, Anders M.~N. Niklasson, Alexander~J. Page, Alessandro Pecchia,
  Gabriele Penazzi, Mats~P. Persson, Jan {\v{R}}ez{\'a}{\v{c}}, Carmen~G.
  S{\'a}nchez, Michael Sternberg, Meike St{\"o}hr, Frank Stuckenberg, Alexandre
  Tkatchenko, Victor W.-z. Yu, and Thomas Frauenheim.
\newblock {DFTB+}, a software package for efficient approximate density
  functional theory based atomistic simulations.
\newblock {\em The Journal of Chemical Physics}, 152(12):124101, 2020.

\bibitem{pearson1897spurious}
Karl Pearson.
\newblock Mathematical contributions to the theory of evolution---on a form of
  spurious correlation which may arise when indices are used in the measurement
  of organs.
\newblock {\em Proceedings of the Royal Society of London}, 60:489--498, 1897.

\bibitem{phipson2010permutation}
Belinda Phipson and Gordon~K. Smyth.
\newblock Permutation p-values should never be zero: Calculating exact p-values
  when permutations are randomly drawn.
\newblock {\em Statistical Applications in Genetics and Molecular Biology},
  9(1):39, 2010.

\bibitem{zheng2021aiqm1}
Peikun Zheng, Roman Zubatyuk, Wei Wu, Olexandr Isayev, and Pavlo~O. Dral.
\newblock Artificial intelligence-enhanced quantum chemical method with broad
  applicability.
\newblock {\em Nature Communications}, 12(1), December 2021.

\bibitem{zeng2023qdpi}
Jinzhe Zeng, Yujun Tao, Timothy~J. Giese, and Darrin~M. York.
\newblock {QD$\pi$}: A quantum deep potential interaction model for drug
  discovery.
\newblock {\em Journal of Chemical Theory and Computation}, 19(4):1261--1275,
  January 2023.

\bibitem{wengert2022cocrystals}
Simon Wengert, Gábor Csányi, Karsten Reuter, and Johannes~T. Margraf.
\newblock A hybrid machine learning approach for structure stability prediction
  in molecular co-crystal screenings.
\newblock {\em Journal of Chemical Theory and Computation}, 18(7):4586--4593,
  June 2022.

\bibitem{mardirossian2016wb97mv}
Narbe Mardirossian and Martin Head-Gordon.
\newblock {$\omega$B97M-V}: A combinatorially optimized, range-separated
  hybrid, meta-{GGA} density functional with {VV10} nonlocal correlation.
\newblock {\em The Journal of Chemical Physics}, 144(21):214110, 2016.

\bibitem{rappoport2010def2svpd}
Dmitrij Rappoport and Filipp Furche.
\newblock Property-optimized {Gaussian} basis sets for molecular response
  calculations.
\newblock {\em The Journal of Chemical Physics}, 133(13):134105, 2010.

\bibitem{parrish2017psi4}
Robert~M. Parrish, Lori~A. Burns, Daniel G.~A. Smith, Andrew~C. Simmonett,
  A.~Eugene DePrince, Edward~G. Hohenstein, U{\u{g}}ur Bozkaya, Alexander~Yu.
  Sokolov, Roberto Di~Remigio, Ryan~M. Richard, J{\'e}r{\^o}me~F. Gonthier,
  Andrew~M. James, Harley~R. McAlexander, Ashutosh Kumar, Masaaki Saitow, Xiao
  Wang, Benjamin~P. Pritchard, Prakash Verma, Henry~F. Schaefer, Konrad
  Patkowski, Rollin~A. King, Edward~F. Valeev, Francesco~A. Evangelista,
  Justin~M. Turney, T.~Daniel Crawford, and C.~David Sherrill.
\newblock {Psi4} 1.1: An open-source electronic structure program emphasizing
  automation, advanced libraries, and interoperability.
\newblock {\em Journal of Chemical Theory and Computation}, 13(7):3185--3197,
  2017.

\bibitem{dftborg2025ptbp}
{DFTB.org}.
\newblock Periodic table baseline parameter set ({PTBP}).
\newblock \url{https://dftb.org/parameters/download.html}, 2025.
\newblock Accessed 26 July 2026.

\bibitem{gameel2026correctedRelease}
Kareem~M. Gameel, Ihor Neporozhnii, Sjoerd Hoogland, and Oleksandr Voznyy.
\newblock When a better baseline makes delta learning worse: separating
  residual scale from learnability to guide target design (code and data
  release).
\newblock
  \url{https://github.com/kareem-gameel/delta-learning-target-design/releases/tag/v0.4.2},
  2026.
\newblock Version v0.4.2.

\bibitem{gameelQM9Psi4Dataset}
Kareem~M. Gameel and Oleksandr Voznyy.
\newblock {Psi4} total electronic energies of 130,807 {QM9} molecules at the
  \ensuremath{\omega}{B97M-V/def2-SVPD} level of theory.
\newblock Zenodo dataset. DOI: \url{https://doi.org/10.5281/zenodo.24203175},
  2026.

\end{thebibliography}

\section*{Acknowledgements}
The authors acknowledge support from the Alliance for AI-Accelerated Materials Discovery (A3MD). Computations were performed on the Trillium supercomputer at the SciNet HPC Consortium. SciNet is funded by Innovation, Science and Economic Development Canada; the Digital Research Alliance of Canada; the Ontario Research Fund: Research Excellence; and the University of Toronto.

\section*{Author contributions}
K.M.G. conceived and designed the study, performed the computational experiments, analysed the data, prepared the figures, and wrote the original manuscript. I.N. contributed to data preparation, computational workflows, and interpretation. O.V. and S.H. supervised the project. O.V. contributed to study design, interpretation, and manuscript revision. All authors reviewed and approved the manuscript.

The authors declare no competing interests.

\end{document}